\documentclass[10pt,twocolumn,letterpaper]{article}

\usepackage[pagenumbers]{cvpr}      %

\usepackage{booktabs} %
\usepackage{graphicx}

\usepackage{multirow}
\usepackage{tabularx}
\newcolumntype{Y}{>{\centering\arraybackslash}X}
\newcolumntype{C}[1]{>{\centering}p{#1}}
\newcolumntype{Z}{>{\raggedleft\arraybackslash}X}
\usepackage{pifont}
\usepackage{xcolor}
\newcommand{\cmark}{\textcolor{green!80!black}{\ding{51}}}
\newcommand{\xmark}{\textcolor{red}{\ding{55}}}

\newcommand*{\hlbest}[1]{\underline{#1}}

\usepackage{multirow}
\newcommand{\STAB}[1]{\begin{tabular}{@{}c@{}}#1\end{tabular}}

\usepackage{xcolor,colortbl} 
\newcommand{\cellgrayed}{-}

\usepackage[outline]{contour}
\newcommand{\uparrowaligned}{\raisebox{0.15em}{\scalebox{0.75}{\contour{black}{$\uparrow$}}}}
\newcommand{\downarrowaligned}{\raisebox{0.15em}{\scalebox{0.75}{\contour{black}{$\downarrow$}}}}

\makeatletter
\renewcommand{\paragraph}{%
  \@startsection{paragraph}{4}%
  {\z@}{0.25ex \@plus 0.25ex \@minus .2ex}{-1em}%
  {\normalfont\normalsize\bfseries}%
}
\makeatother

\definecolor{cvprblue}{rgb}{0.21,0.49,0.74}
\usepackage[pagebackref,breaklinks,colorlinks,allcolors=cvprblue]{hyperref}

\title{\vspace{-1em}Charge: A Comprehensive Novel View Synthesis \\ Benchmark and Dataset to Bind Them All }
\author{Michal Nazarczuk \quad Thomas Tanay \quad Arthur Moreau \quad Zhensong Zhang \quad Eduardo Pérez-Pellitero\\
Huawei Noah's Ark Lab \\
London, UK\\
{\tt\small [michal.nazarczuk1, thomas.tanay, arthur.moreau3, zhangzhensong, e.perez.pellitero]@huawei.com}
}

\begin{document}

\twocolumn[{%
\renewcommand\twocolumn[1][]{#1}%
\maketitle
\centering
\vspace{-2em}
\includegraphics[width=0.7\linewidth]{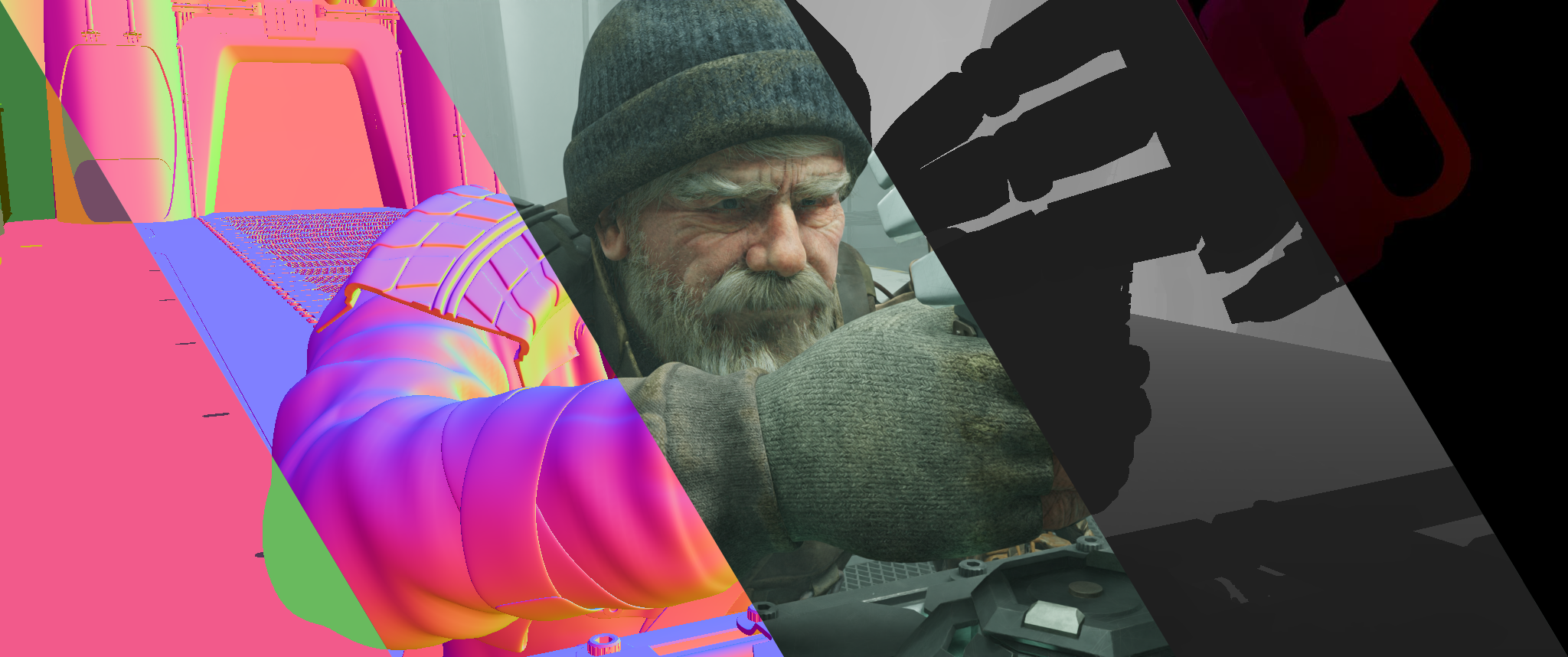}
\captionof{figure}{Charge is a new dataset that binds diverse scenarios related to static and dynamic novel view synthesis, characterised by high-quality renderings and rich annotations. The example includes left-to-right: object segmentation, normals, RGB, depth, and optical flow.
\vspace{1em}
}
\label{fig:teaser}
}]

\begin{abstract}
This paper presents a new dataset for Novel View Synthesis, generated from a high-quality, animated film with stunning realism and intricate detail. Our dataset captures a variety of dynamic scenes, complete with detailed textures, lighting, and motion, making it ideal for training and evaluating cutting-edge 4D scene reconstruction and novel view generation models. In addition to high-fidelity RGB images, we provide multiple complementary modalities, including depth, surface normals, object segmentation and optical flow, enabling a deeper understanding of scene geometry and motion. The dataset is organised into three distinct benchmarking scenarios: a dense multi-view camera setup, a sparse camera arrangement, and monocular video sequences, enabling a wide range of experimentation and comparison across varying levels of data sparsity. With its combination of visual richness, high-quality annotations, and diverse experimental setups, this dataset offers a unique resource for pushing the boundaries of view synthesis and 3D vision.
\end{abstract}
\vspace{-1em}
\section{Introduction} \label{sec:intro}
Access and availability of high-quality, large-scale datasets for training and evaluation of algorithms is one of the main engines for progress and innovation in computer vision. With the advent of the internet, an abundance of readily available images unlocked unprecedented developments in many tasks, \eg image classification, or low-level vision, especially when leveraging forms of self-supervision that alleviate the need for costly manual data annotation. 

\begin{table*}
\centering
\caption{A summary and comparison of datasets used in dynamic novel view synthesis. The top section includes datasets used for multi-view evaluation whereas the middle section focuses on monocular evaluation data. $^{\dagger}$ - 2 camera rig with alternating frames assigned to training and test trajectory.
}
\renewcommand{\arraystretch}{0.9}
\begin{tabular}{lccccccccc}  %
    {\small Dataset} & {\small Dense} & {\small Sparse} & {\small Mono} & {\small Depth} & {\small FPS} &  {\small \#\,Seq} & {\small \#\,Train\,cams} & {\small \#\,Test\,cams} &  {\small \#\,Frames} \\ %
    \midrule
    Neural 3D Video\,\cite{li2022a} & \cmark & \xmark & \xmark & \xmark & 30 & 6 & 20 & 1 & 56\,700 \\ %
    Technicolor\,\cite{sabater2017} & \cmark & \xmark & \xmark & \xmark & 30 & 5 & 15 & 1 & 25\,696 \\ %
    Google Immersive\,\cite{broxton2020} & \cmark & \xmark & \xmark & \xmark & 30 & 7 & 45 & 1 & 157\,320\\
    \midrule
    Nvidia Dynamic\,\cite{yoon2020novel} & \cmark & \xmark & \xmark & \cmark & 60 & 8 & 1 & 11 & 2\,304\\ %
    D-NeRF\,\cite{pumarola2020} & \xmark &  \xmark & \cmark & \xmark & 60 & 8 & 1 & 1 & 1\,410 \\
    Nerfies\,\cite{park2021b} & \xmark &  \xmark & \cmark & \xmark & 5 & 4 & 1$^{\dagger}$ & 1$^{\dagger}$ & 1\,680 \\
    HyperNeRF\,\cite{park2021a} & \xmark &  \xmark & \cmark & \xmark & 15 & 17 & 1$^{\dagger}$ & 1$^{\dagger}$ & 2\,152 \\
    DyCheck\,\cite{gao2022} & \xmark &  \xmark & \cmark & \cmark & 60 & 7+7 & 1 & 2/0  & 8\,746\\
    \midrule
    Ours & \cmark & \cmark & \cmark & \cmark &  96 & 8 & 25+9+4 & 16+10+16 & 185\,600 \\
    \bottomrule
\end{tabular}
\label{tab:related_work}
\vspace{-1em}
\end{table*}

Recently, NeRF\,\cite{mildenhall2020} or 3DGS\,\cite{kerbl2023} reconstruct the scene by fitting a representation supervised via differentiable rendering and photometric loss on a set of posed 2D images. This coupled with \textit{off-the-shelf} Structure-from-Motion (SfM) camera pose estimators such as COLMAP~\cite{schoenberger2016sfm}, can reconstruct 3D from a handful of 2D images. 
Differentiable rendering has received a significant number of follow-up works, \eg dense, sparse, varying appearance. Most of these assume static scenes and fixed radiance fields. 

A new generation of feed-forward foundational models, pioneered by VGGT\,\cite{wang2025}, has emerged as a unified architecture capable of predicting multiple 3D representations (camera poses, depth, point clouds, 3D Gaussians \cite{jiang2025}) in a single forward pass. These models promise efficient inference and broad generalisation across diverse scenarios without per-scene optimisation. Due to lack of better options, these foundational models have been evaluated almost exclusively on unimodal benchmarks (\ie separately for camera pose estimation, depth, NVS), leaving a gap in the evaluation to further understand and assess the relationships between the output modalities, and more generally between geometry and appearance.

Extending static to dynamic scene reconstruction is also one of the most promising directions of research, either as a feed-forward or optimisation methodologies. There has been recent interest in addressing this problem, with early paradigms extending NeRF~\cite{pumarola2020, park2021a} or 3DGS~\cite{li2024, wu2024, zhu2024motiongs} with temporal deformation. Capturing dynamic scenes is inherently difficult. As opposed to static scenes, camera pose estimation on dynamic scenes is a challenging and largely unsolved problem, \ie movement in the scene can introduce matching errors that then degrade the estimated poses. Furthermore, due to low image sampling rates (\eg 24 fps), \textit{soft} or inaccurate camera synchronisation can in turn impact the camera pose and reconstruction accuracy. 
In practice, complex, hardware-synchronised multi-camera systems are needed to obtain multiple views of the scene~\cite{lu2024diva,li2022_neural3d}. Such rigs restrict the scene diversity due to the limitations on physical spaces where they can be deployed (\ie indoors, volumetric studios), and ultimately can not be moved within the scene (in part due to weight, fragility and build, but also because of potential problems with camera calibration). 
The need for testing views results in a reduction of the number of training views available, a trade-off that often result in poor camera sampling for evaluation. For approaches where separated, non-synchronised cameras are used for the testing views~\cite{gao2022}, problems can arise in the camera calibration in dynamic scenes. This, coupled with the lack of precise temporal frame alignment, degrades the ability to accurately evaluate the performance of 4D reconstruction methods.

\begin{figure*}
    \centering
    \includegraphics[width=0.8\linewidth]{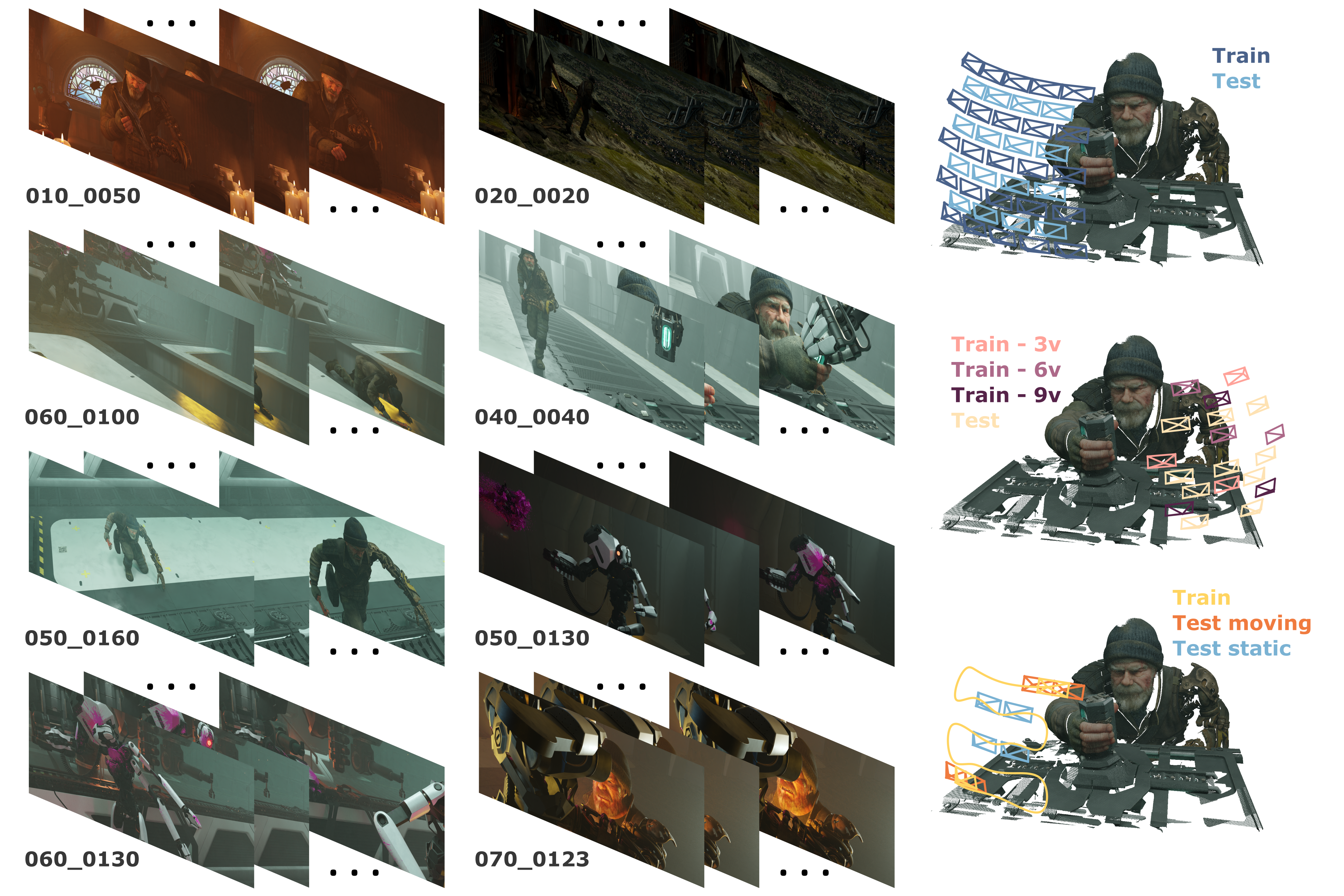}
    \caption{An overview of the Charge dataset. The left side presents a selection of frames from all the animations. On the right side, a 3 setups included in the dataset are presented: Dense, Sparse, and Mono. Cameras allocation and sample movement path for monocular trajectory are overlayed on the section of the point cloud corresponding to scene 040\_0040.}
    \label{fig:overview}
    \vspace{-1em}
\end{figure*}

Given the technical limitations to capture 4D posed data, we find inspiration in the \textit{MPI Sintel} dataset~\cite{butler2012}. In our work, we present a novel high-resolution, dynamic dataset for unified 3D evaluation with rich modalities (RGB, depth, normals, segmentation), and overcomplete training and testing camera coverage that addresses open challenges of 3D reconstruction. The dataset is arranged for different static and dynamic reconstruction setups: dense , sparse, and monocular inputs, and supports full evaluation of current and next generation foundational models (camera pose, depth, NVS). We show a visualisation of the modalities in Fig.\,\ref{fig:teaser}, and an overview of the dataset composition in Fig.\,~\ref{fig:overview}. This dataset is generated from the high-quality animated movie Charge\,\cite{charge} including rich scene compositions and a variety of content. The synthetic nature of the dataset allows us to introduce a large-scale dataset and benchmark not viable to be collected in the real world due to capture cost, physical limitations, and ultimately accurate pose estimation. We summarise our contributions as follows:
\begin{enumerate}
    \item We introduce a new synthetic dataset - Charge - for next generation models benchmarking, including dynamic NVS, 3D foundational models. It is characterised by high visual quality, it outscales other available datasets in the number of images, and it provides rich annotations offering a variety of additional modalities.
    \item We unify all relevant reconstruction setups - Static, Dynamic, Dense, Sparse, Mono, and provide benchmarking for all of them.
    \item We perform an extensive evaluation of state-of-the-art reconstruction methods and analyse the results emphasising the importance of the benchmark we propose.
\end{enumerate}

\section{Related Work} \label{sec:rw}

Advances in novel view synthesis include the introduction of two seminal reconstruction paradigms, namely Neural Radiance Fields \cite{mildenhall2020} and 3D Gaussian Splatting\,\cite{kerbl2023}. These developments in static scene reconstruction were quickly followed by several works on dynamic content.

\noindent
\textbf{Dynamic Reconstruction:}
NeRF-based methods for video reconstruction include D-NeRF\,\cite{pumarola2020}, StreamRF\,\cite{li2022b}, HexPlane\,\cite{cao2023}, K-Planes\,\cite{fridovich2023}, Tensor4D\,\cite{shao2023}, MixVoxels~\cite{wang2023}. Representative monocular approaches include NSFF\,\cite{li2020neural}, Nerfies\,\cite{park2021b}, HyperNeRF\,\cite{park2021a}, DyCheck \cite{gao2022}.
Similarly, Gaussian Splatting enabled reconstruction of multi-view videos, \eg GaussianFlow\,\cite{lin2024}, 4DGS\,\cite{wu2024}, STG\,\cite{li2024}, MotionGS\,\cite{zhu2024motiongs}, SWinGS \cite{shaw2024}, Ex4DGS\,\cite{lee2024}.
The most recent advances in Gaussian Splatting include monocular video reconstructions by works such as Deformable 3D Gaussians\,\cite{yang2024}, SC-GS\,\cite{huang2024}, or MoSca\,\cite{lei2024}.

\noindent
\textbf{Foundational models:}
The recent advances in feed-forward geometry estimation emphasise the need for unified benchmarking. VGGT\,\cite{wang2025} introduced a transformer architecture for multiple 3D tasks, including camera parameter estimation, depth estimation, point cloud reconstruction, 3D tracking. This, along with follow-up works \cite{keetha2025, wang2024, wang2025b} established a group of foundational 3D models trained in a supervised manner on large-scale data. Several works offer pipelines that predict gaussians from a set of unposed views \cite{zhang2025, hong2025, huang2025}. AnySplat\,\cite{jiang2025} and WorldMirror\,\cite{liu2025} build upon VGGT to enable gaussian splat prediction for NVS.

\noindent
\textbf{Datasets:} Current evaluation of 
dynamic NVS is based on a handful of datasets summarised in Table\,\ref{tab:related_work}. This includes real-world data. Neural 3D Video\,\cite{li2022a} introduces indoor cooking scenes captured with a static rig. Technicolor\,\cite{sabater2017} provides several scenes captured with a rig of $4\!\!\times\!\!4$ cameras. The shortcoming of the two datasets is the reduced amount of dynamic content (\eg cooking occupies only a small region of the videos), and relatively slow motions. Additionally, those datasets evaluate the performance only on $1$ camera selected from the rig. Google Immersive\,\cite{broxton2020} provides videos captured with a spherical rig of outward-facing cameras and suffers from heavy fish-eye distortions and very low overlap between views. DiVa-360\,\cite{lu2024diva} captures a set of objects and hand-object interactions. Stereo4D\,\cite{jin2025} offers a dataset extracted from stereoscopic fisheye videos. DynPose\,\cite{rockwell2025} filters and annotates internet videos with camera poses. 
Monocular datasets include \eg Nvidia Dynamic Scene\,\cite{yoon2020novel} which introduces sequences captured by a rig of 12 cameras subsampling alternating views to create a monocular trajectory. Nerfies\,\cite{park2021b} and HyperNeRF\,\cite{park2021a} capture action with two cameras and alternate their frames in the input and test sequences. They suffer from highly unrealistic motion due to teleporting cameras creating thus a pseudo-multi-view capture. This is characterised as a high effective multi-view factor in DyCheck\,\cite{gao2022}, which introduces a new monocular capture with $2$ static cameras for evaluation. However, it suffers from an imperfect set of camera poses. A better set of camera poses can be seen in DTU\,\cite{aanaes2016} captured with a camera mounted on a robotic arm. It is broadly used in novel view synthesis evaluation in dense and sparse setups. However, it provides only static scenes. Similarly, Sparse Neural Rendering Dataset\,\cite{nazarczuk2024a} provides a DTU-like setup but in a synthetic version, and thus is also static and present a single object per scene.
D-NeRF\,\cite{pumarola2020} is a synthetic dataset of dynamic content. It captures lower-quality subject assets on a white background.
Further, it creates a monocular input sequence by sampling consecutive cameras (suffering from teleporting-related issues). 
In contrast, Charge provides high-quality renderings of rich sequences with carefully designed setups for dense and sparse multi-view, and monocular captures.

\begin{figure}
    \centering
    \includegraphics[width=0.76\linewidth]{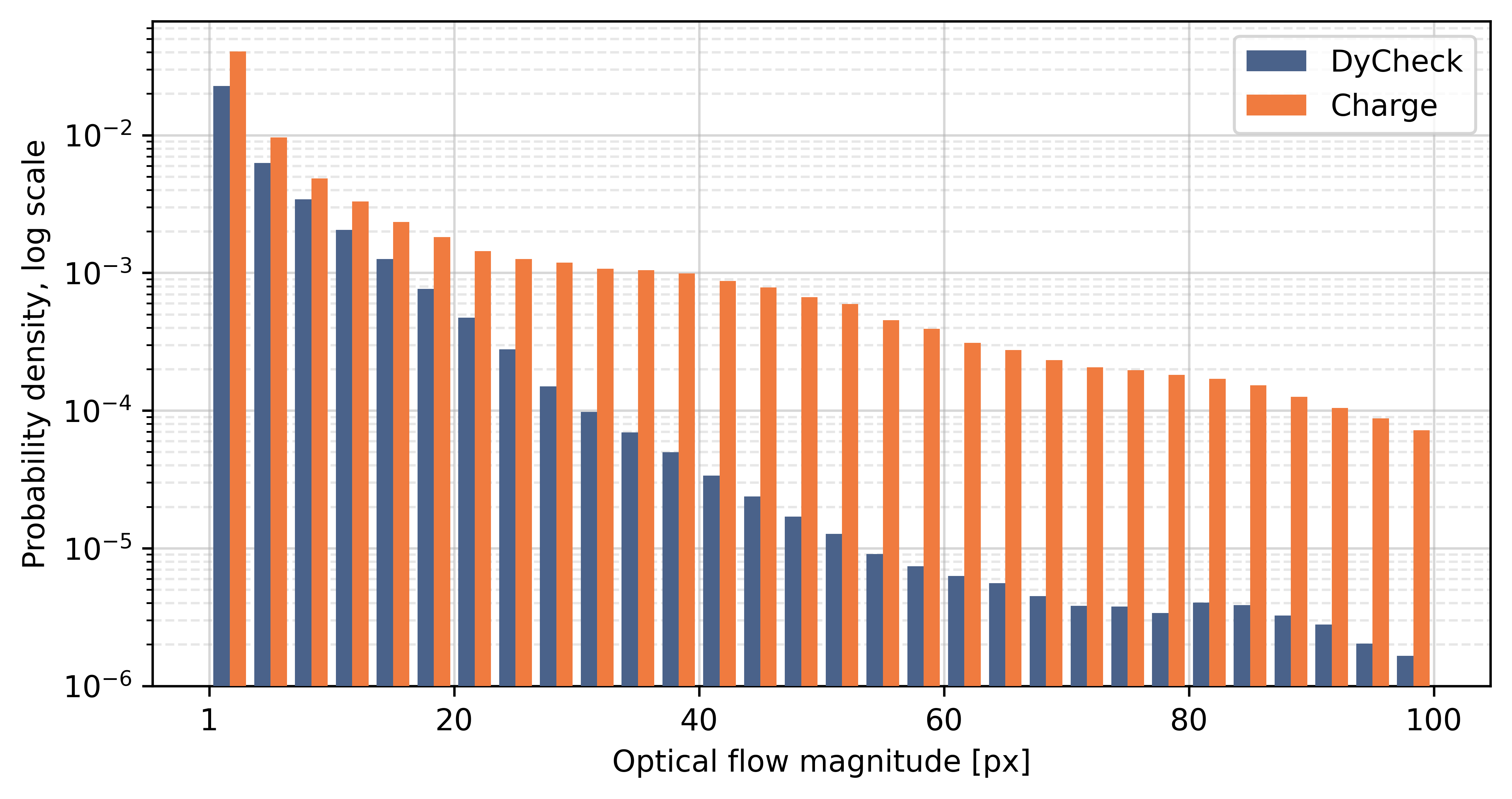}
    \vspace{-3mm}
    \caption{Optical flow histogram for DyCheck and Charge.}
    \label{fig:of_histo}
\vspace{-1em}
\end{figure}

In Table\,\ref{tab:dynamic_content}, we show a breakdown of dynamic content in current datasets by presenting a percentage of pixels occupied by dynamic objects. Notably, frames in Charge are occupied on average by twice the amount of dynamic parts than the closest competitor. Similarly, in Figure~\ref{fig:of_histo} we show histograms of optical flow magnitude for our Charge in comparison to DyCheck\,\cite{gao2022} for their respective dynamic regions. Charge has higher density of dynamic pixels, and has better coverage across the diverse magnitudes, with special advantages on large motions.

\begin{table}
\centering
\small
\caption{Percentage of dynamic content in various datasets.}
\vspace{-2mm}
\label{tab:dynamic_content}
\begin{tabular}{lrrrr}
    \toprule
        Data & {\small Charge} & {\small DyCheck} & {\small Neural 3D} & {\small Technicolor}\\
    \midrule
         Dynamic & $25.1\,\%$ & $12.6\,\%$ & $10.9\,\%$ & $9.7\,\%$ \\
    \bottomrule
\end{tabular}
\vspace{-1em}
\end{table}

\section{The Charge Dataset}

In this work we introduce a new dataset for evaluating new generation models, including but not limited to dynamic NVS and 3D foundation models. In pursue of high-quality data, we explored animated movies created in the rendering software Blender\,\cite{blender}. Specifically, we focus on Charge, a short movie set in a dystopian world, created to showcase Blender's photorealistic rendering capabilities.

\subsection{Dataset Creation}

We utilise production shots 
from the Charge movie to create our dataset. For each such scene a composition with animation, lighting, and a library of items is available. We processed all $8$ available scenes for Charge of which examples can be seen in Figure\,\ref{fig:overview} (left). For every scene, we manually picked places for cameras for the best capture and visibility of the scene after spanning the whole rig. Additionally, we removed the original rendering pipeline due to postprocessing effects that may degrade the 3D consistency of the capture. Instead, we replace the rendering pipeline with direct rendering from the camera, and we add rendering of additional modalities (such as depth, or normals). 

\subsection{Camera Setups}

The Charge dataset provides the unique benchmarking feature of having various camera setups corresponding to commonly tackled novel view synthesis problems within the exact same scenes, namely: Dense, Sparse, and Mono. Commonly used camera setups include the use of coplanar capture or dome capture. Inspired by that, we decided to place the cameras on the section of the sphere, such that scenes are mostly forward-facing (similar to most real-world captures), yet offer the benefits of changing perspective. To this end, we manually adjust the position of the rig and the radius of the aforementioned sphere for each scene.

\textbf{Dense} setup consists of $25$ training cameras and $16$ testing cameras (see Fig.\,\ref{fig:overview} top right). This offers a setup similar to Neural 3D Video\,\cite{li2022a} or Technicolor\,\cite{sabater2017}. Those works, limited by the constraints of the real world (high capture cost), offer only one camera for testing. We provide $16$ testing cameras allowing for a more thorough evaluation.

\begin{table*}
\centering
\caption{Quantitative evaluation results of Charge dataset - Dense and Mono setups, -D denotes metrics on dynamic-only areas, -S static only and $FOV_O$ quantifies the field-of-view overlap between testing and training views (\ie harder when lower). \hlbest{Best} performer.
}
\setlength{\tabcolsep}{5pt}
\renewcommand{\arraystretch}{0.85}
\footnotesize
\begin{tabularx}{\linewidth}{lZZZZZZZZ}
    \toprule
        Method & PNSR & PNSR-D & PNSR-S & SSIM & SSIM-D & LPIPS & LPIPS-D & $FOV_O$\\
    \midrule
        \multicolumn{9}{c}{\textbf{Dense}} \\
    \midrule
        4DGS \cite{wu2024} & $28.94$ & $26.84$ & \hlbest{$31.85$} & $0.881$ & $0.848$ & $0.231$ & $0.307$ & $0.70$ \\
        STG \cite{li2024} & $29.29$ & $28.15$ & $31.33$ & $0.886$ & \hlbest{$0.866$} & $0.193$ & \hlbest{$0.263$} & $0.70$ \\
        Ex4DGS \cite{lee2024} &  \hlbest{$29.75$} & \hlbest{$28.57$} & $31.57$ & \hlbest{$0.893$} & $0.863$ & \hlbest{$0.187$} & $0.264$ & $0.70$ \\
    \midrule
        \multicolumn{9}{c}{\textbf{Mono - Spline Fast}} \\
    \midrule
        4DGS \cite{wu2024} & $24.85$ & $22.69$ & \hlbest{$27.25$} & $0.840$ & $0.794$ & $0.264$ & $0.352$ & $0.42$ \\
        D-3DGS \cite{yang2024} & \hlbest{$24.88$} & $22.91$ & $27.24$ & \hlbest{$0.869$} & $0.807$ & \hlbest{$0.201$} & $0.295$ & $0.42$ \\
        SC-GS \cite{huang2024} & $23.97$ & \hlbest{$23.57$} & $25.89$ & $0.847$ & $0.798$ & $0.217$ & \hlbest{$0.291$} & $0.42$ \\
        MoSca \cite{lei2024} & $23.39$ & $22.84$ & $24.10$ & $0.853$ & \hlbest{$0.817$} & $0.255$ & $0.320$ & $0.42$ \\
    \midrule
        \multicolumn{9}{c}{\textbf{Mono - Spline Slow}} \\
    \midrule
        4DGS \cite{wu2024} & \hlbest{$24.18$} & $21.95$ & \hlbest{$26.65$} & $0.834$ & $0.787$ & $0.266$ & $0.360$ & $0.41$ \\
        D-3DGS \cite{yang2024} & \hlbest{$24.18$} & $22.53$ & $26.15$ & \hlbest{$0.858$} & $0.794$ & \hlbest{$0.203$} & \hlbest{$0.290$} & $0.41$ \\
        SC-GS \cite{huang2024} & $23.29$ & $22.54$ & $25.15$ & $0.843$ & $0.786$ & $0.225$ & $0.304$ & $0.41$ \\
        MoSca \cite{lei2024} & $23.45$ & \hlbest{$22.83$} & $24.63$ & $0.856$ & \hlbest{$0.815$} & $0.244$ & $0.323$ & $0.41$ \\
    \midrule
        \multicolumn{9}{c}{\textbf{Mono - Random Walk Fast}} \\
    \midrule
        4DGS \cite{wu2024} & \hlbest{$24.78$} & $22.79$ & \hlbest{$26.81$} & $0.835$ & $0.790$ & $0.267$ & $0.359$ & $0.41$ \\
        D-3DGS \cite{yang2024} & $24.34$ & $22.55$ & $26.24$ & $0.852$ & $0.789$ & \hlbest{$0.217$} & \hlbest{$0.309$} & $0.41$ \\
        SC-GS \cite{huang2024} & $22.90$ & \hlbest{$23.04$} & $24.22$ & $0.822$ & $0.775$ & $0.243$ & $0.314$ & $0.41$ \\
        MoSca \cite{lei2024} & $23.82$ & $22.77$ & $25.09$ & \hlbest{$0.857$} & \hlbest{$0.817$} & $0.251$ & $0.325$ & $0.41$ \\
    \midrule
        \multicolumn{9}{c}{\textbf{Mono - Random Walk Slow}} \\
    \midrule
        4DGS \cite{wu2024} & $23.38$ & $21.68$ & \hlbest{$25.64$} & $0.818$ & $0.771$ & $0.277$ & $0.370$ & $0.38$ \\
        D-3DGS \cite{yang2024} & $22.85$ & $21.61$ & $24.76$ & $0.829$ & $0.762$ & \hlbest{$0.227$} & $0.318$ & $0.38$ \\
        SC-GS \cite{huang2024} & $21.86$ & $21.73$ & $23.51$ & $0.811$ & $0.762$ & $0.244$ & $0.318$ & $0.38$ \\
        MoSca \cite{lei2024} & \hlbest{$24.29$} & \hlbest{$23.61$} & $25.23$ & \hlbest{$0.871$} & \hlbest{$0.829$} & $0.229$ & \hlbest{$0.306$} & $0.38$ \\
        \bottomrule
\end{tabularx}
\label{tab:results_dense_mono}
\vspace{-1em}
\end{table*}

\textbf{Sparse} setup follows a common practise in literature (\eg  pixelNeRF\,\cite{yu2021}) and we provide training scenarios including $3$, $6$, or $9$ training cameras, accompanied by $10$ testing views (see Fig.\,\ref{fig:overview} middle right). Note that the sparse setup covers the scenes from a different perspective instead of simply subsampling the training cameras. This increases the diversity of our dataset and opens up new areas of research and exploration when performing  distant novel view synthesis, \ie with an emphasis on view extrapolation as opposed to  interpolation.

\textbf{Mono} setup introduces $4$ different trajectories of the monocular camera. We aim to reproduce in this setup camera motions that are viable with a handheld device. As characterised by DyCheck, a monocular capture should not exhibit a high effective multiview factor (\eg teleporting or extremely fast camera). To this end, based on empirical experimentation with the use of a phone camera and a target captured from around $1-2\text{m}$, we estimate that the reasonable velocity of a capture device lies around the range of $15-50\,\text{cm/s}$. Therefore, we introduce \textit{Fast} and \textit{Slow} monocular trajectories corresponding roughly to the upper and lower bounds of the measured values. Similarly, we propose that the capture may deliberately try to cover a wider perspective or be less regularised. Hence, we suggest two qualities of the camera trajectory: \textit{Spline} - constrained to a spline curve 
spanning the majority of corresponding training camera positions, and \textit{Random Walk} - randomly generated direction for every time step with a smoothing factor. This results in the following training scenarios: \textit{SplineFast}, \textit{SplineSlow}, \textit{RandomWalkFast}, \textit{RandomWalkSlow}. Further, monocular reconstruction encompasses many downstream goals, such as stabilisation, or spatial video (stereo pair synthesis). Thus, we propose to test each of these trajectories with $4$ static cameras chosen from the set of Dense testing cameras (allowing for direct comparison), and $4$ cameras defined relative to the training one (different baselines, and a camera orbiting around).

\subsection{Data Qualities}

As visible on the left-hand side of Figure\,\ref{fig:overview}, Charge introduces a large diversity of scenes posing challenging and interesting problems for 3D reconstruction methods. Examples include a scene with large open bounds and a smaller dynamic subject (020\_0020), with a large amount of movement (040\_0040), with a large area of dynamic content (070\_0123), or highly non-rigid elements (050\_0130). 

With the development of 3DGS, the high-resolution evaluation became viable and is arguably preferable. Thus, we render our dataset in a resolution of \textbf{2048$\times$858}. Further, we provide a high-speed capture in \textbf{96 \text{fps}}. Additionally, thanks to the synthetic nature of our data, we are able to provide highly precise camera poses in contrast to real-world captures that rely on estimations. Finally, along with \textit{RGB images}, we supply \textbf{metric depth, optical flow, normals, UV maps, object segmentation, and dynamic content masks}.

In total, Charge offers \textbf{8} diverse scenes, each rendered with \textbf{25+16} dense setup cameras, \textbf{9+10} sparse setup cameras, and \textbf{4+16} monocular setup cameras. This accounts for a total of \textbf{185\,600} frames available, providing a larger scale of data than any other existing
datasets (see Tab.\,\ref{tab:related_work}).

\section{Benchmark}

We evaluate the Charge dataset with a selection of methods representing the state-of-the-art in multi-view and monocular Gaussian Splatting reconstruction. We report a large selection of metrics, introducing a measure of difficulty for each task, and analyse the results, focusing on challenges available in our dataset that emphasise problems that are difficult for current methods to deal with. We include dynamic evaluation on scene-specific methods and a static evaluation on next generation foundational models.

\begin{table*}
\centering
\caption{Quantitative evaluation results of Charge dataset - Sparse setup, -D denotes metrics on dynamic-only areas, -S static only and $FOV_O$ quantifies the field-of-view overlap between testing and training views (\ie harder when lower). \hlbest{Best} performer.
}
\setlength{\tabcolsep}{5pt}
\renewcommand{\arraystretch}{0.85}
\footnotesize
\begin{tabularx}{\linewidth}{lrZZZZZZZZ}
    \toprule
        Method & Views & PNSR & PNSR-D & PNSR-S & SSIM & SSIM-D & LPIPS & LPIPS-D & $FOV_O$\\
    \midrule
        \multicolumn{10}{c}{\textbf{Sparse}} \\
    \midrule
        {4DGS \cite{wu2024}} &  & \hlbest{$19.71$} & \hlbest{$20.71$} & \hlbest{$20.28$} & \hlbest{$0.776$} & \hlbest{$0.740$} & $0.358$ & $0.416$ & $0.54$ \\
        {STG \cite{li2024}} & \textbf{3} & $18.80$ & $20.15$ & $19.04$ & $0.753$ & $0.721$ & $0.371$ & $0.418$ & $0.54$ \\
        {Ex4DGS \cite{lee2024}} &  & $19.53$ & $20.34$ & $20.05$ & $0.759$ & $0.717$ & \hlbest{$0.350$} & \hlbest{$0.400$} & $0.54$ \\
    \midrule
        {4DGS \cite{wu2024}} &  & \hlbest{$23.93$} & \hlbest{$24.89$} & \hlbest{$24.29$} & \hlbest{$0.840$} & \hlbest{$0.810$} & \hlbest{$0.277$} & \hlbest{$0.338$} & $0.62$ \\
        {STG \cite{li2024}} & \textbf{6} & $22.39$ & $23.61$ & $22.57$ & $0.820$ & $0.790$ & $0.295$ & $0.361$ & $0.62$ \\
        {Ex4DGS \cite{lee2024}} &  & $22.07$ & $23.30$ & $22.31$ & $0.813$ & $0.781$ & $0.299$ & $0.348$ & $0.62$ \\
    \midrule
        {4DGS \cite{wu2024}} & & \hlbest{$26.67$} & \hlbest{$26.51$} & \hlbest{$27.45$} & \hlbest{$0.874$} & \hlbest{$0.840$} & $0.226$ & $0.301$ & $0.64$ \\
        {STG \cite{li2024}} & \textbf{9} & $24.52$ & $24.96$ & $25.03$ & $0.850$ & $0.815$ & $0.260$ & $0.347$ & $0.64$ \\
        {Ex4DGS \cite{lee2024}} &  & $24.50$ & $26.26$ & $24.73$ & $0.859$ & $0.829$ & \hlbest{$0.225$} & \hlbest{$0.282$} & $0.64$ \\
    \bottomrule
\end{tabularx}
\label{tab:results_sparse}
\vspace{-1em}
\end{table*}

\subsection{Dynamic Benchmark}

We provide evaluation for all $3$ setups in our dataset, namely Dense, Sparse, and Mono. Therefore, we use 4D Gaussian Splatting\,\cite{wu2024}  to train all sequences in all setups, as it is a method suitable for both multi- and single-view reconstruction. Further, we use Spacetime Gaussians\,\cite{li2024} and Ex4DGS\,\cite{lee2024} to evaluate Dense and Sparse scenarios. Finally, in a monocular setting, we add 3 dedicated methods, Deformable 3D Gaussians\,\cite{yang2024}, SC-GS\,\cite{huang2024}, and MoSca\,\cite{lei2024}. All methods were trained on full-length sequences and evaluated in full resolution of data in Charge. %

\paragraph{Metrics}
In our evaluation, we report commonly used metrics, such as PSNR, SSIM\,\cite{wang2004}, and LPIPS\,\cite{zhang2018}. Further, for PSNR on dynamic benchmark, we include the metric calculated within the mask of dynamic and static content (PSNR-D and PSNR-S, respectively). Similarly, for SSIM and LPIPS we report a version of these metrics calculated for dynamic content based on a tight bounding box of the dynamic mask (SSIM-D, LPIPS-D). 

Additionally, we propose to quantify the difficulty of each task. To this end, we suggest calculating a parameter indicating an overlap between the field of views of the testing camera with respect to training cameras. To this end, for each testing view, we reproject the image plane 
into all training views obtaining coverage masks - $\{m_i\}$ (effectively saying which area of the testing view is visible from the given training view). Further, we sum up all such masks and normalise them by the number of pixels in training views $n_{train}\cdot HW$, obtaining a weighted covisibility proxy (where pixels seen by more training views have a value closer to $1$). Therefore, averaging for $n_{test}$ test views, the field of view overlap is expressed as:
\begin{equation}
    FOV_O = \frac{1}{n_{test}} \left( \frac{\sum_{n_{train}} m_i}{n_{train} \cdot HW} \right) 
\end{equation}
Finally, we average the value for all the test views to obtain an indication of task difficulty. 
A more detailed analysis, along with the breakdown of the metrics, including separation per scene is available in the Supplementary Material.

\begin{figure}
    \centering
    \includegraphics[width=0.9\linewidth]{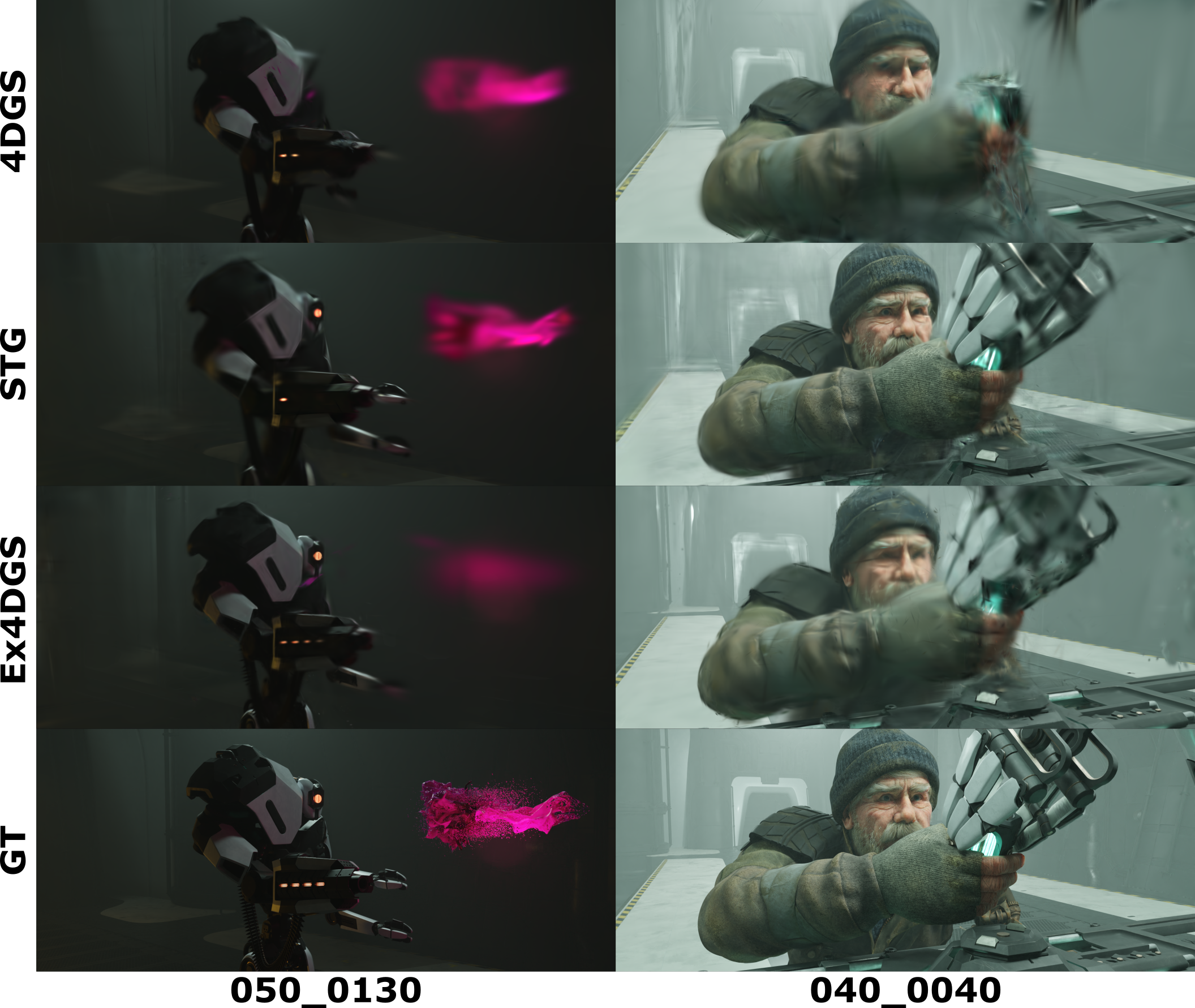}
    \caption{Example results of rendering in Charge dataset evaluation - Dense setup. 
    Best viewed zoomed in.}
    \label{fig:results_dense}
    \vspace{-1em}
\end{figure}

\begin{figure*}
    \centering
    \includegraphics[width=0.82\linewidth]{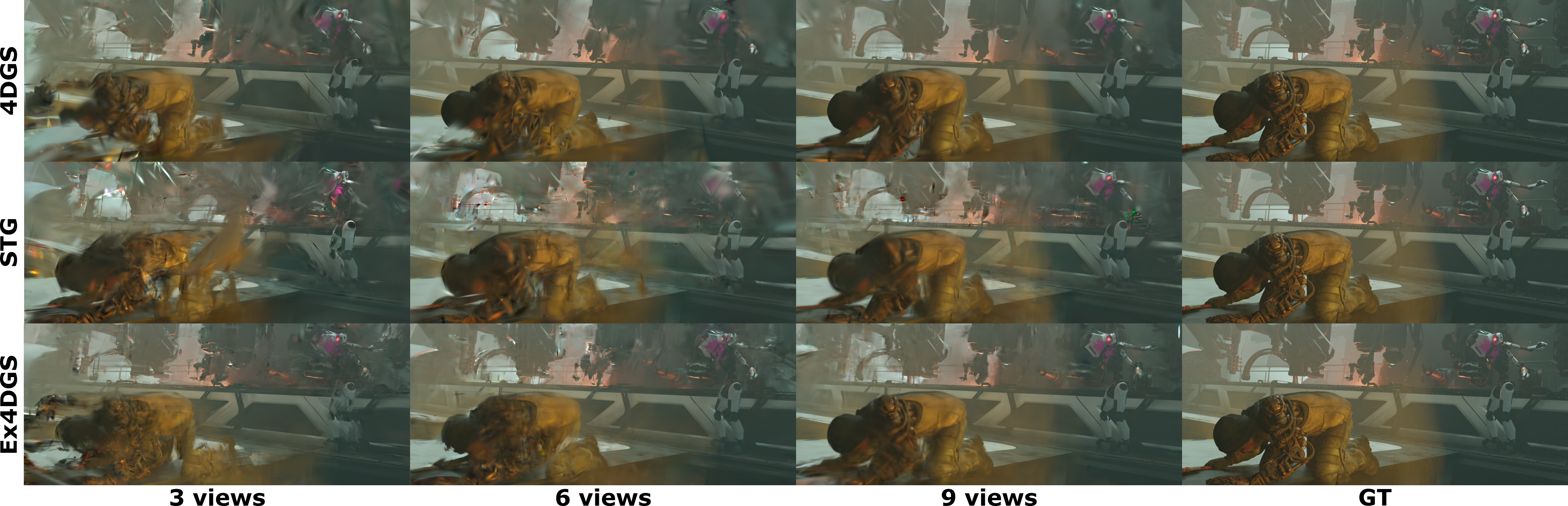}
    \caption{Example results of rendering in Charge dataset evaluation - Sparse setup, scene 050\_0160. Best viewed zoomed in.}
    \label{fig:results_sparse}
    \vspace{-1em}
\end{figure*}

\begin{figure}
    \centering
    \includegraphics[width=0.89\linewidth]{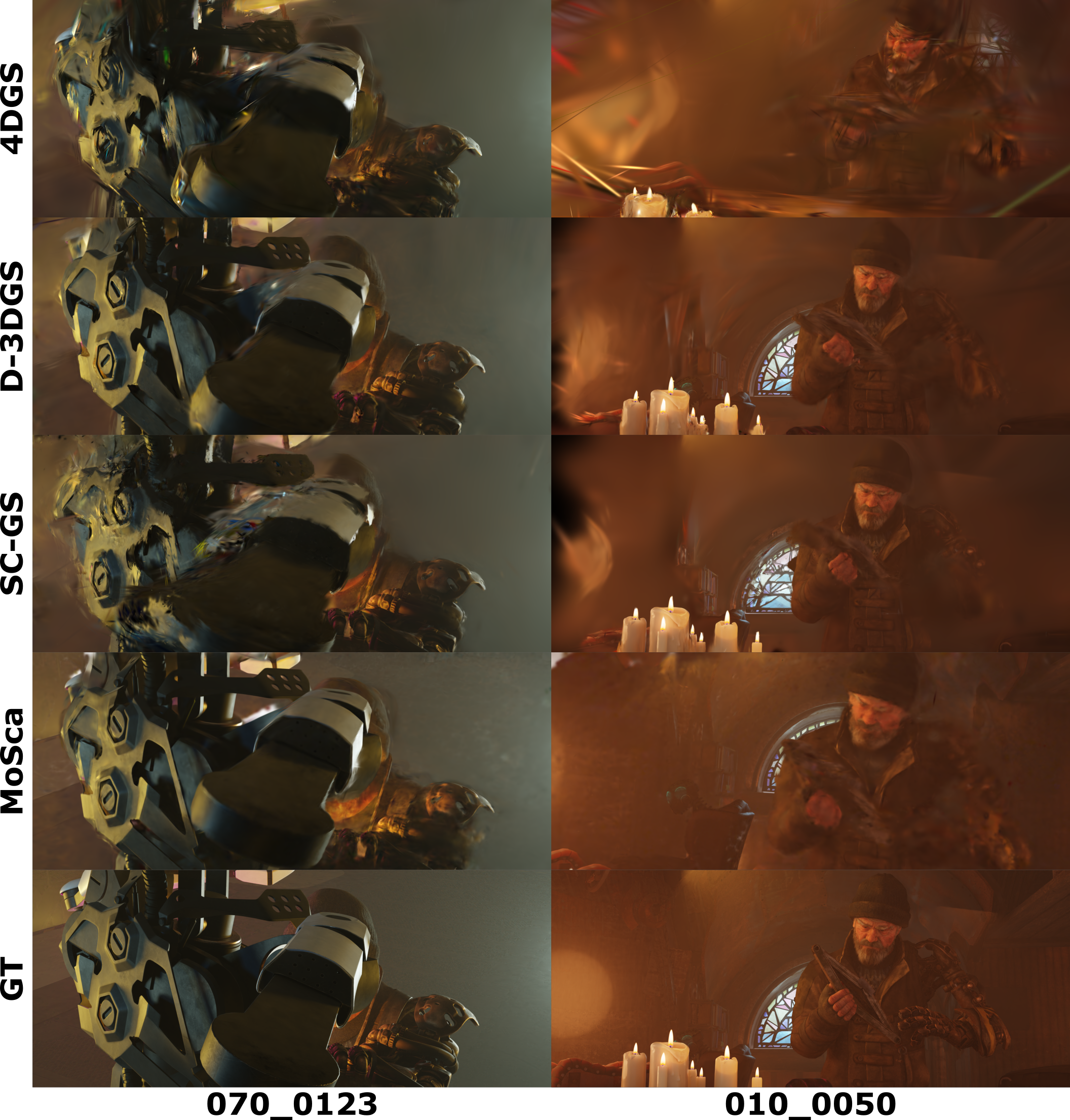}
    \caption{Example results of rendering in Charge dataset evaluation - Mono setup. Best viewed zoomed in.}
    \label{fig:results_mono}
    \vspace{-1em}
\end{figure}

\paragraph{Dense}
The results of the experiments on the Dense setup of the Charge dataset are presented in Table\,\ref{tab:results_dense_mono} (top rows). Additionally, selected qualitative samples can be seen in Figure\,\ref{fig:results_dense}. We can observe PSNR results of $28.94$, $29.29$, and $29.75$. This indicates that Charge poses a good challenge for current state-of-the-art methods. In Fig.\,\ref{fig:results_dense} (left side) we can see that all methods struggle with reconstructing a splash of paint which leads us to believe that even in the dense multi-view camera setup, the reconstruction of the dynamics of a highly non-rigid object is yet to be explored in depth. 
Further, in Fig.\,\ref{fig:results_dense}, right, 
we note that our dataset provides a good way of evaluating the effect of the length of the sequence on the reconstruction (with sequences ranging from 93 to 653 frames). The given example shows that 4DGS struggles with reconstructing a long-range sequence with a large amount of movement. 
Notably, however, we do not notice a decrease in the performance with the consecutive frames but rather a consistently lower quality of the reconstruction. Finally, unsurprisingly, we can observe the gap between the reconstruction of dynamic and static regions which is reflected in the presented results. This leads us to believe that data like Charge with a large amount of dynamic content is crucial for an insightful evaluation.

\paragraph{Sparse}
In Table\,\ref{tab:results_sparse} we present the results of the evaluation on the Sparse setup of the Charge dataset. Note that as indicated in Figure\,\ref{fig:overview} (right), the sparse camera setup does not overlap with the dense one. Therefore, the corresponding numerical values should not be directly compared. However, based on the values of the field of view overlap, we notice that sparse rendering is significantly harder than dense reconstruction. Additionally, we note that the difficulty increase from $6$ to $3$ training views is higher than from $9$ to $6$ training views. An example selection of scenes trained and rendered in all scenarios 
is shown in Figure\,\ref{fig:results_sparse}. We can observe that in the lower camera regime, 4DGS and Ex4DGS perform slightly better than STG, indicating that the selection of the best method for the task is configuration-dependent. Further, we can clearly observe, both visually, and in the presented metrics, a drastic difference in performance with respect to the number of training cameras. In Figure\,\ref{fig:results_sparse} we can see that even in $9$-views case, the renderings exhibit a lack of fine details (\eg hands in the example). Further, with $6$ training views, we notice an increasing number of background missing or incorrect gaussian. Finally, $3$ views present a very hard case for the benchmarked methods, leading to very artefact-filled renders. Interestingly, the differences in the reconstruction of dynamic and static content are not as prominent in the Sparse setup. 
We believe that the decrease in the overlap between cameras leads to higher uncertainties, especially in disoccluded areas, as well as edge areas, seen by a very low number of cameras.

\begin{table*}
\centering
\caption{A static benchmark of next generation foundational models on Charge, including camera pose estimation, depth estimation and novel view synthesis, \textsuperscript{\textdagger} denotes target views used for camera alignment. \hlbest{Best} performer.
}
\setlength{\tabcolsep}{4pt}
\renewcommand{\arraystretch}{0.9}
\footnotesize
\begin{tabularx}{\linewidth}{clZZZZZZZZZZZ}
    \toprule
        & \multirow{2.4}{*}{Method} & \multicolumn{3}{c}{Camera pose} & \multicolumn{2}{c}{Depth} & \multicolumn{3}{c}{NVS} & \multicolumn{3}{c}{NVS\textsuperscript{\textdagger}}\\
    \cmidrule(lr){3-5}\cmidrule(lr){6-7}\cmidrule(l){8-10}\cmidrule(lr){11-13}
        &  & {\scriptsize AUC}@5\,\uparrowaligned{} & {\scriptsize AUC}@15\,\uparrowaligned{} & {\scriptsize AUC}@30\,\uparrowaligned{} & AbsRel\,\downarrowaligned{} & {\footnotesize $\delta \!\! < \!\! 1.25$}\,\uparrowaligned{} & PSNR\,\uparrowaligned{} & SSIM\,\uparrowaligned{} & LPIPS\,\downarrowaligned{} & PSNR\,\uparrowaligned{} & SSIM\,\uparrowaligned{} & LPIPS\,\downarrowaligned{} \\
    \midrule
        \multirow{4}{*}{\STAB{\rotatebox[origin=c]{90}{Sparse-3}}} & {VGGT\,\cite{wang2025}} & \hlbest{0.3581} & \hlbest{0.7599} & \hlbest{0.8809} & 0.1039 & 0.8669 & \cellgrayed{} & \cellgrayed{} & \cellgrayed{} & \cellgrayed{} & \cellgrayed{} & \cellgrayed{} \\
        & {$\pi^3$\,\cite{wang2025b}} & 0.1900 & 0.6376 & 0.8127 & \hlbest{0.1014} & \hlbest{0.8895} & \cellgrayed{} & \cellgrayed{} & \cellgrayed{} & \cellgrayed{} & \cellgrayed{} & \cellgrayed{} \\
        & {AnySplat\,\cite{jiang2025}} & 0.2827 & 0.7394 & 0.8708 & 0.2101 & 0.7557 & 12.15 & 0.6723 & 0.6080 & 13.34 & 0.7065 & 0.5656 \\
        & {WorldMirror\,\cite{liu2025}} & 0.3547 & 0.7148 & 0.8555 & 0.2003 & 0.7565 & \hlbest{17.54} & \hlbest{0.7318} & \hlbest{0.5648} & \hlbest{20.94} & \hlbest{0.7841} & \hlbest{0.5110} \\
    \midrule
        \multirow{4}{*}{\STAB{\rotatebox[origin=c]{90}{Sparse-6}}}  & {VGGT\,\cite{wang2025}} & 0.5009 & 0.8086 & 0.9046 & 0.0959 & 0.8679 & \cellgrayed{} & \cellgrayed{} & \cellgrayed{} & \cellgrayed{} & \cellgrayed{} & \cellgrayed{} \\
        & {$\pi^3$\,\cite{wang2025b}} & 0.2844 & 0.7118 & 0.8565 & \hlbest{0.0896} & \hlbest{0.9191} & \cellgrayed{} & \cellgrayed{} & \cellgrayed{} & \cellgrayed{} & \cellgrayed{} & \cellgrayed{} \\
        & {AnySplat\,\cite{jiang2025}} & 0.3751 & 0.7733 & 0.8871 & 0.1721 & 0.7970 & 13.27 & 0.6972 & 0.5822 & 15.95 & 0.7359 & 0.5441 \\
        & {WorldMirror\,\cite{liu2025}} & \hlbest{0.5049} & \hlbest{0.8164} & \hlbest{0.9087} & 0.1601 & 0.7873 & \hlbest{18.97} & \hlbest{0.7560} & \hlbest{0.5472} & \hlbest{21.86} & \hlbest{0.7958} & \hlbest{0.4997} \\
    \midrule
        \multirow{4}{*}{\STAB{\rotatebox[origin=c]{90}{Sparse-9}}}  & {VGGT\,\cite{wang2025}} & \hlbest{0.5188} & \hlbest{0.8161} & \hlbest{0.9087} & 0.0992 & 0.8681 & \cellgrayed{} & \cellgrayed{} & \cellgrayed{} & \cellgrayed{} & \cellgrayed{} & \cellgrayed{} \\
        & {$\pi^3$\,\cite{wang2025b}} & 0.2886 & 0.7183 & 0.8596 & \hlbest{0.0909} & \hlbest{0.9133} & \cellgrayed{} & \cellgrayed{} & \cellgrayed{} & \cellgrayed{} & \cellgrayed{} & \cellgrayed{} \\
        & {AnySplat\,\cite{jiang2025}} & 0.3731 & 0.7667 & 0.8837 & 0.1981 & 0.7727 & 16.30 & 0.7345 & 0.5541 & 19.52 & 0.7755 & 0.5101 \\
        & {WorldMirror\,\cite{liu2025}} & 0.4919 & 0.8066 & 0.9032 & 0.1781 & 0.7904 & \hlbest{19.57} & \hlbest{0.7671} & \hlbest{0.5436} & \hlbest{22.84} & \hlbest{0.8122} & \hlbest{0.4956} \\
    \midrule
        \multirow{4}{*}{\STAB{\rotatebox[origin=c]{90}{Dense}}}  & {VGGT\,\cite{wang2025}} & \hlbest{0.5149} & 0.8011 & 0.8996 & \hlbest{0.1085} & 0.8499 & \cellgrayed{} & \cellgrayed{} & \cellgrayed{} & \cellgrayed{} & \cellgrayed{} & \cellgrayed{} \\
        & {$\pi^3$\,\cite{wang2025b}} & 0.4071 & 0.7634 & 0.8807 & 0.1344 & \hlbest{0.8893} & \cellgrayed{} & \cellgrayed{} & \cellgrayed{} & \cellgrayed{} & \cellgrayed{} & \cellgrayed{} \\
        & {AnySplat\,\cite{jiang2025}} & 0.3359 & 0.7524 & 0.8767 & 0.2248 & 0.7595 & 18.74 & 0.7440 & 0.5533 & 23.12 & 0.7943 & \hlbest{0.4955} \\
        & {WorldMirror\,\cite{liu2025}} & 0.5126 & \hlbest{0.8205} & \hlbest{0.9093} & 0.1811 & 0.7632 & \hlbest{20.10} & \hlbest{0.7666} & \hlbest{0.5475} & \hlbest{24.44} & \hlbest{0.8192} & 0.4962 \\
    \bottomrule
\end{tabularx}
\label{tab:results_static}
\vspace{-1em}
\end{table*}

\paragraph{Mono}
We propose $4$ configurations for evaluating monocular trajectories and provide evaluation separated as such, namely \textit{Spline Fast}, \textit{Spline Slow}, \textit{Random Walk Fast}, \textit{Random Walk Slow}. The results from all configurations are presented in Table\,\ref{tab:results_dense_mono} (bottom rows). A selection of qualitative results for the Mono setup is shown in Fig.\,\ref{fig:results_mono}. This setup is evaluated based on $8$ cameras - $4$ cameras correspond to central testing cameras from the Dense setup, and $4$ further cameras are moving together with the training one ($3$ represent different baselines in the task of creating stereo pairs, and $1$ camera is orbiting around the training one).

Unsurprisingly, we can notice a decrease in the field of view overlap (increase in difficulty) for this task when compared to the dense setup. We see slight differences between the configurations of the Mono setup, indicating structured (spline covering the area better) and faster (more area covered) motion to be easier to reconstruct. 

We notice a similar performance of 4DGS and D-3DGS, and lower for SC-GS.
Surprisingly, MoSca (best on DyCheck\cite{gao2022}), performs the best only in the structural metric, being lower on PSNR. 
This may be caused by evaluation on our dataset with a large amount of motion. 
MoSca heavily relies on trajectory estimation in a preprocessing step, which seems to be harder on Charge due to the larger range of motion and objects moving into the camera view. Additionally, MoSca optimises camera poses during training, and relies on test time optimisation to update test poses, introducing uncertainty with respect to underlying reconstruction. This emphasises the need for Charge benchmark allowing evaluation with perfect ground-truth cameras.
In Fig.\,\ref{fig:results_mono} we notice that when reconstructing a scene with a large amount of dynamic content close to the camera (\eg 070\_0123), the methods struggle with the reconstruction of such regions. Similarly, as seen in 010\_0050, we notice how monocular reconstruction is affected by a lack of coverage outside the field of view causing dark areas near the edges.

\subsection{Static Benchmark}

In addition, we subselect a set of frames from each scene and propose a static benchmark to evaluate foundational models which currently focus on static scenes only. To this end, we use our \textbf{Sparse} and \textbf{Dense} setup and evaluate camera pose estimation, depth estimation, and novel view synthesis. We evaluate VGGT\,\cite{wang2025} and $\pi^3$\,\cite{wang2025b} that do not predict the appearance of the scene, as well as AnySplat\,\cite{jiang2025} and WorldMirror\,\cite{liu2025} that produce Gaussian splats. We present all results in Table\,\ref{tab:results_static}.

\paragraph{Camera pose estimation} We report AUC at angular thersholds of 5$^{\circ}$, 15$^{\circ}$, and 30$^{\circ}$. AUC is the area under the accuracy-threshold curve of the minimum values between Relative Rotation Accuracy and Relative Translation Accuracy across varying thresholds. We note a close performance between VGGT and WorldMirror which share a VGGT backbone. Notably AnySplat even though initialised from VGGT performs worse indicating a degradation of camera head performance in the model. $\pi^3$ is noticably worse in lower threshold metrics.

\paragraph{Depth estimation} We report Absolute Relative Error and threshold accuracy of 25$\%$. In this task, $\pi^3$ is the best performer, in contrast to camera pose evaluation. VGGT's performance is a close second, whereas the methods that output Gaussian splats decrease the depth estimation performance in the course of photometric supervision.

\paragraph{Novel View Synthesis} We report PSNR, SSIM, LPIPS, a commonly used metrics in novel view synthesis. The benchmark methods provide results in an arbitrary scale. Therefore, we distinguish two testing protocols in this task present in literature. In first, we align the target cameras to the models prediction by fitting a ground truth to predicted pose transformation on the source views (Umeyama\,\cite{umeyama1991}). In the second, we follow AnySplat protocol, \ie we run the model again on the set of both source and target views to then use the predicted target cameras to render Gaussians predicted from source views. In this experiment we observe WorldMirror outperforming AnySplat. This correlates with camera pose estimation results, and in qualitative examples (Supp.\,Mat.), we observe misalignment between target and ground truth images. Further, the Umeyama alignment performs significantly worse than using target views for alignment. Given the perfect ground truth camera pose, this emphasises a pose-shape ambiguity in the predicted geometries. 
Importantly, when including photometric consistency in the pipeline, the methods preserve the relative ranking across all tasks in contrast to directly supervised ones (VGGT and $\pi^3$). This suggest that the coupling of NVS and geometry tasks may have a negative effect on the performance of the current models.
Notably, this analysis can only be performed in a dataset with perfect ground truth for both geometry and appearance, such as Charge.

\section{Conclusions}

In this paper, we introduced Charge, a new dataset and benchmark for novel view synthesis. Our dataset was rendered from a professionally created animation movie. This enables us to obtain photorealistic data, with detailed shapes and textures as well as intricate movements and interactions. Charge offers several advantages over real-world datasets: (1) various capture setups for different problems (static, dynamic, dense, sparse, monocular), (2) with high number of testing cameras, and (3) with accurate camera poses, thus increasing the reliability of the evaluation. (4)~Charge contains more motion than any other dataset to date, and (5) this motion is more diverse, \ie ranging from small to large displacements. (6) Charge includes rich annotations, including $6$ data modalities, encouraging the development of new methods and novel evaluation paradigms.

We extensively tested the performance of SoTA GS-based and 3D foundational models on this data. We provided a thorough analysis of the selected methods in all proposed setups. Our evaluation emphasises the shortcomings of current approaches by 
showing areas that need improvement, based on several aspects and input features, \eg camera speed, trajectory style, static vs dynamic performance. %
Finally, the performance of the state-of-the-art methods in the Charge benchmark shows that the data we propose is challenging, and that it can be an accelerator in developing and evaluating the next generation NVS approaches.

{
    \small
    \bibliographystyle{ieeenat_fullname}
    \bibliography{main}
}

\end{document}